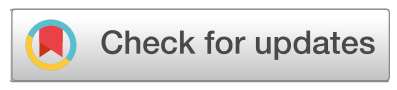

OPEN

# A comprehensive interpretable machine learning framework for mild cognitive impairment and Alzheimer's disease diagnosis

Maria Eleftheria Vlontzou[1]✉, Maria Athanasiou[1], Kalliopi V. Dalakleidi[1], Ioanna Skampardoni[1], Christos Davatzikos[2,3] & Konstantina Nikita[1]

**An interpretable machine learning (ML) framework is introduced to enhance the diagnosis of Mild Cognitive Impairment (MCI) and Alzheimer's disease (AD) by ensuring robustness of the ML models' interpretations. The dataset used comprises volumetric measurements from brain MRI and genetic data from healthy individuals and patients with MCI/AD, obtained through the Alzheimer's Disease Neuroimaging Initiative. The existing class imbalance is addressed by an ensemble learning approach, while various attribution-based and counterfactual-based interpretability methods are leveraged towards producing diverse explanations related to the pathophysiology of MCI/AD. A unification method combining SHAP with counterfactual explanations assesses the interpretability techniques' robustness. The best performing model yielded 87.5% balanced accuracy and 90.8% F1-score. The attribution-based interpretability methods highlighted significant volumetric and genetic features related to MCI/AD risk. The unification method provided useful insights regarding those features' necessity and sufficiency, further showcasing their significance in MCI/AD diagnosis.**

Dementia is a term used to describe several diseases that damage the brain and affect memory, thinking, and the ability to perform daily activities, and is currently the seventh leading cause of death[1]. Alzheimer's Disease (AD) is the most common form of Dementia, representing 60-70% of Dementia cases. The strongest known risk factor for AD is age. Additional risk factors include hypertension, diabetes, obesity, smoking, social isolation, physical inactivity, and depression[1]. Mild Cognitive Impairment (MCI) is a condition related to memory and thinking problems and is a risk factor for developing AD. Clinical Decision Support Systems (CDSSs) are increasingly making use of Artificial Intelligence (AI) to speed up MCI or AD diagnosis and optimise disease management. AI models have been shown to exceed the accuracy of radiologists' predictions, as they can effectively use the rich information present in dense, heterogeneous and high-dimensional data[2]. They also show promise at identifying those at risk earlier in the disease trajectory, when longitudinal clinical observations are available[2].

However, the integration of AI in clinical decision-making entails a spectrum of challenges, tied to requirements that stem from various algorithmic and data-related aspects, such as the heterogeneity of healthcare data, drawn from diverse sources and, the presence of class imbalance, which may introduce bias, thus undermining the models' generalisation abilities[3]. Furthermore, the imperative need for transparency and interpretability in AI models presents a persistent hurdle as complex algorithms must yield comprehensible, trustworthy outputs to facilitate informed decision-making among healthcare practitioners. All of these issues are particularly pronounced in the context of MCI or AD diagnosis, which involves the consideration of multifaceted data, often characterised by stark class imbalance, and requires the generation of comprehensive insights, underlying diagnostic recommendations provided to healthcare professionals and patients.

To address these challenges, several approaches utilising Machine Learning (ML) for MCI or AD diagnosis have been proposed, either focusing on the classification of patients with AD with respect to healthy controls or undertaking the task of multi-class classification, aiming at distinguishing among individuals with MCI, healthy controls, and AD patients. In the case of the classification of AD patients with respect to healthy controls, the proposed models have exhibited varying performance in terms of accuracy, ranging from 77.0% to 98.8%. A multi-modal approach combining a fully Convolutional Neural Network (CNN) CNN with age, gender, and cognitive scores, as well as an ensemble of 3D densely connected CNN, utilising 3D brain Magnetic Resonance

[1]Faculty of Electrical and Computer Engineering, National Technical University of Athens, Athens 15773, Greece. [2]Center for Biomedical Image Computing and Analytics, University of Pennsylvania, Philadelphia, PA, USA. [3]Department ofRadiology, University of Pennsylvania, Philadelphia, PA, USA. ✉email: mvlontzou@biosim.ntua.gr

Imaging (MRI), have achieved the highest discrimination performance[4,5]. Several studies have performed multi-class classification to classify patients with varying degrees of disease severity[2], while a couple of studies have investigated the potential to discriminate patients with MCI who later converted to AD (progressive MCI (pMCI)) from those that remained stable (stable MCI (sMCI)). In this case, the reported classification accuracy has varied between 65.4% and 88.5%, which could be attributed to the ambiguous definition of the MCI label across medical healthcare providers and the absence of distinct neuropathological differences between these two groups[2].

It is noteworthy that despite the promising performance of these studies, research until now has mostly focused on those models' predictive power, rather than the understanding of their predictions and behaviour. Another obstacle is that, in the effort to improve the predictive accuracy of ML algorithms, their complexity has increased, leading to an additional difficulty in interpreting their predictions. The continuously emerging need for fair, trustworthy, robust, and highly efficient ML models has led to the flourishing of eXplainable Artificial Intelligence (XAI), which entails the application of interpretability methods for producing explanations on the systems' behaviour and final output[6]. When it comes to CDSS for MCI/AD diagnosis, interpretability methods utilising visualisation techniques or feature ranking based on learned weights have been applied to enhance the models' transparency[7]. Moreover, the SHapley Additive exPlanations (SHAP) interpretability method has been utilised towards the identification of highly influential features for the models' predictions[6,8–11].

In studies of non-imaging data, including electronic health records and clinical data, well-established risk factors such as age, smoking, cardiovascular problems, and lack of exercise were indicated as predictive of future AD diagnosis[2]. For studies with neuroimaging data, interpretability methods involved overlaying heatmaps on brain scans that highlight the regions associated with a specific class. For example, the classification model in[4] identified the brain regions of temporal lobes, hippocampus, cingulate cortex, corpus callosum, and parts of the parietal and frontal lobes as important for classifying a brain scan to an AD patient. With respect to studies utilising multimodal data, such as demographic data, mental state exams, volumetric measurements from brain imaging, and genetic data, the Rey Auditory Verbal Learning Test was identified in[12] as a distinguishing feature even in the presence of other imaging-derived features. The authors of[13] also included this test as a feature but found that the Alzheimer's Disease Assessment Scale (ADAS) and Functional Activities Questionnaire were more useful. Given that cognitive tests are designed specifically to be used as AD biomarkers, the produced interpretations, highlighting the importance of these tests, were in accordance with existing clinical knowledge.

In terms of the robustness of the produced interpretations, although some studies have recently focused on quantifying the robustness of specific types of interpretability methods separately[14–16], or measuring the stability of feature attribution rankings provided by XAI methods[17], limited research has been dedicated to the investigation of approaches for unifying different interpretability methods and assessing their robustness altogether. In this direction, Mothilal et al.[18] have developed a method based on actual causality, which evaluates the necessity and sufficiency of the models' features by combining attribution-based and counterfactual-based interpretability methods. In the case of AD, no study has investigated the unification of multiple interpretability methods towards the development of interpretable prediction models for MCI or AD diagnosis, able to provide robust and reliable models interpretations.

To address the need for reliable, transparent, and trustworthy decision support tools in AD management, the present study introduces a comprehensive methodological framework towards the development of interpretable ML models for MCI or AD diagnosis. The main contribution of the proposed framework is threefold:

- Firstly, it employs a method based on ensemble learning for one versus one classification, able to address both multiclass and imbalanced data problems, by engaging and thoroughly evaluating several machine learning classification algorithms for MCI and AD diagnosis.
- Additionally, it leverages a combination of MRI volumetric measurements of 145 anatomical brain Regions of Interest (ROIs), together with 54 AD related Single Nucleotide Polymorphisms (SNPs) towards the diagnosis of both MCI and AD, while also jointly measuring and comparing the contribution of features from those modalities in patients' classification.
- Lastly, it employs a plurality of interpretability methods, including attribution-based and counterfactual-based approaches, in order to provide human-friendly explanations for the model's predictions and increase its reliability, by capturing various aspects of how each feature affects and contributes to MCI and AD diagnosis, while providing information related to the clinical and physiological traits of the subjects. Most importantly, it focuses on the unification of the different interpretability techniques by utilising a previously proposed method based on actual causality[18], which calculates the features' necessity and sufficiency. To the best of our knowledge, this is the first time these metrics are employed towards assessing the robustness of interpretability methods in the context of MCI and AD diagnosis.

The proposed framework is analysed in the "Methods" section and a schematic representation of its components is depicted in Fig. 1.

## Results

### Classifiers' discrimination performance

Various classifiers based on the use of Random Forests (RFs), Logistic Regression (LR), Multi Layer Perceptron (MLP), Support Vector Machines (SVMs), Gradient Boosting (GB), and eXtreme Gradient Boosting (XGBoost), were trained and evaluated within the One versus All (OVA), One versus One (OVO), and Bagging with OVO decomposition schemes, which were explored to address the multiclass and imbalanced aspects of the problem. Table 1 summarises the results from the classifiers' performance evaluation. The weighted F1-score and balanced accuracy metrics are compared, as obtained by (1) an 80:20 Train Test Split (TTS) and (2) the Bagging method after the hyperparameters' tuning based on the application of a $5 \times 4$ fold nested cross validation scheme. The last three columns refer to the mean, the maximum, and the standard deviation values of the two metrics for the best model of each fold in the outer 5-fold cross validation.

**Fig. 1**. Schematic overview of the proposed conceptual framework, which comprises the classification methods, the evaluation framework, the interpretability methods and the framework for unifying interpretations.

| | One vs All TTS | | One vs One TTS | | Bagging TTS | | Bagging CV (Mean) | | Bagging CV (Max) | | Bagging CV (STD) | |
|---|---|---|---|---|---|---|---|---|---|---|---|---|
| | F1-score | Bal. Acc. | F1-score | Bal. Acc. | F1-score | Bal. Acc. | F1-score | Bal. Acc. | F1-score | Bal. Acc. | F1-score | Bal. Acc. |
| RF | 54.4% | 47.9% | 75.1% | 72.1% | 88.8% | 84.3% | 90.6% | 86.9% | 92.9% | 88.4% | 1.4% | 1.4% |
| LR | 53.6% | 51.1% | 77.1% | 76.9% | 77.3% | 78.7% | 88.3% | 85.3% | 90.9% | 88.3% | 1.6% | 1.8% |
| MLP | 50.5% | 47.2% | 76.5% | 75.0% | 81.2% | 82.9% | 85.2% | 83.3% | 86.3% | 84.5% | 1.2% | 1.0% |
| SVM | 53.8% | 48.5% | 78.0% | 76.1% | 89.9% | 86.1% | **90.8%** | **87.5%** | 92.6% | 89.6% | 2.3% | 1.9% |
| GB | 55.9% | 50.8% | 79.2% | 76.7% | 86.9% | 84.6% | 88.2% | 85.6% | 93.1% | 90.2% | 2.6% | 2.8% |
| XGB | 53.9% | 48.3% | 80.7% | 78.7% | 86.7% | 84.2% | 89.6% | 86.9% | 93.8% | 90.8% | 2.4% | 2.3% |

**Table 1**. Comparison of the different classification methods (One versus All, One versus One and Bagging with One versus One before and after the hyperparameter tuning of the cross fold validation), which were applied with six different classifiers (Random Forest, Logistic Regression, Support Vector Machines, Multilayer Perceptron, Gradient Boosting, XGBoost) by means of the weighted F1-score and the balanced accuracy metrics. The optimal metrics obtained by the highest-performing model are highlighted in bold. TTS: train test split, CV: cross fold validation.

In general, the OVO approach achieved higher performance compared to OVA, but the use of the Bagging ensemble method, including the OVO decomposition scheme, yielded even higher classification performance. After the hyperparameters' tuning of the ensemble classifiers trained using the Bagging method, the performance metrics showed a small increase as presented in Table 1. Out of the six ensemble classifiers the highest mean value of balanced accuracy was provided by the SVM and the next best classifiers were the RF and the XGBoost, based on the mean value obtained from the cross validation. The application of the statistical t-test revealed no statistically significant differences among the classifiers' performance metrics (p-values > 0.05), with the only exception being the MLP, which exhibited an overall significantly lower performance (RF-MLP: p-value = 0.0026, SVM-MLP: p-value = 0.0050, XGBoost-MLP: p-value = 0.0201).

## Explanations from individual interpretability methods

Meaningful insights were obtained by applying various interpretability methods, including the Gini index[19], SHAP[20], Local Interpretable Model-agnostic Explanations (LIME)[21], Partial Dependence Plots (PDPs)[22], and counterfactual explanations, to the two best performing classifiers, i.e., the SVMs and the RF. For brevity, the interpretability results presented here only refer to the MCI versus AD binary subproblem, since those two classes were the most challenging to be distinguished from one another. The results from the classification of the other two binary subproblems were similar and can be found in the Supplementary material.

Because of their tree-based structure, the predictions of the RF, the GB, and the XGBoost classifiers were interpreted by providing impurity-based feature importances as described by the Gini measure. In order to obtain the most important features utilised by the tree-based classifiers, those classifiers with the optimal hyperparameters were applied separately to each pair of classes. When distinguishing between the MCI and AD classes with RF, the most prevailing features were the right inferior temporal gyrus, the left lateral ventricle, the left hippocampus, the left inferior temporal gyrus and the right middle temporal gyrus. Notably, many of these features also appeared in the top positions of the SHAP feature ranking, as described in the following subsection.

More details about the features highlighted by the Gini importance method can be found in the Supplementary material.

In order to obtain the feature importance ranking of the SVM classifier, the SHAP framework was utilised. By calculating the mean absolute SHAP values of each feature, the features were ordered in decreasing order of importance. Figure 2a depicts the feature ranking provided by the classification of the MCI and AD classes. The most important features included, among others, the right and left lateral ventricles, the right entorhinal area, and the left and right middle temporal gyrus. For the exploration of the attribution of the features values, the SHAP method was applied separately on the three binary classification problems to produce SHAP Summary plots. Figure 2b depicts the Summary plot obtained by the classification of MCI and AD patients with the SVM Classifier, where the features are ranked in descending order of importance. It is noteworthy that most of the highly ranked features according to SHAP were also characterised as highly important by the Gini index. It can be observed that a high volume of the right entorhinal area or other features, such as the left and right middle temporal gyrus, contributed to an individual being classified in the "negative" class (i.e., MCI according to the MCI versus AD problem). Low values of the same features were indicative of brain atrophy and contributed to the "positive" (AD) class. Additionally, a lower volume of the left and right lateral ventricle indicated an individual belonging to the "negative" class, while a higher volume showed that an individual was more likely to be an AD patient ("positive" class).

The LIME method was also applied separately for each of the implemented classifiers and the studied binary subproblems. Local explanations were obtained for indicative prediction cases, including a True Positive, a True Negative, a False Positive, and a False Negative randomly selected instance. For the MCI versus AD problem, MCI patients were considered as the "negative" samples and AD patients as the "positive" ones. In the instance depicted in Fig. 3, the ROI volumes that increased a sample's probability of belonging to the AD rather than the MCI class, included, among others, the right fusiform gyrus, the left hippocampus, the right middle temporal gyrus, and the right inferior temporal gyrus. These explanations aligned with the results obtained from the SHAP global feature ranking for the same pair of classes. Conversely, the features that increased the sample's probability of belonging to the MCI class were, among others, the right and left entorhinal areas and the right and left lateral ventricles. Figure 3 shows the explanations derived from the LIME method for an AD patient who was misclassified as an MCI patient by the SVM-based classifier.

The use of PDPs was investigated during the classification with SVMs to study the dependence of the prediction outcome on selected features. Apart from the features (mostly ROIs) which were highlighted by the SHAP method, the importance of additional individual features, including certain SNPs, was also explored. The obtained results demonstrated that specific SNPs were found to enhance an individual's probability of having been diagnosed with MCI or AD, whereas the presence of others seemed to lower this probability (as shown in the Supplementary material). For one of the most highly ranked features, namely the right lateral ventricle, as seen in Fig. 4a, an increase in its volume influenced the AD prediction positively and vice versa. The presence

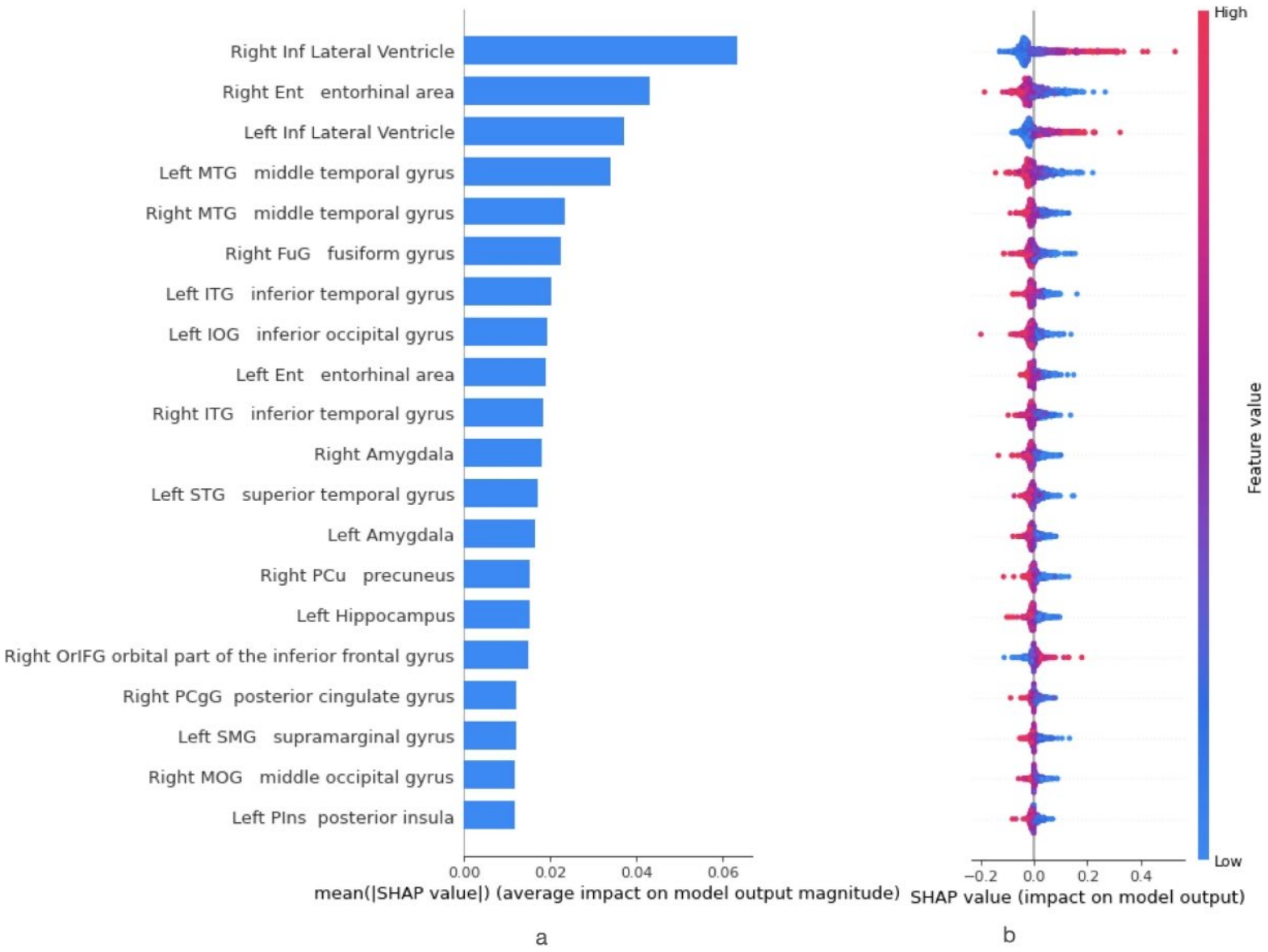

**Fig. 2**. (**a**) Feature Importance ranking for the classification of the Mild Cognitive Impairment (MCI) and Alzheimer's disease (AD) classes by the Support Vector Machines model based on the calculation of the mean absolute SHAP values. (**b**) SHAP method Summary plot for the classification of the Mild Cognitive Impairment (MCI) classes and Alzheimer's disease (AD) classes with a Support Vector Classifier. Features are ranked based on their importance values. Every point represents the SHAP value of an instance. Red points correspond to high feature values, whereas blue points indicate low values. When a data point is located on the left (right) of the y-axis, it represents an instance with a negative (positive) SHAP value for the specific feature, reflecting this feature's contribution to the instance being classified in the negative—MCI (positive—AD) class.

Prediction probabilities

| | |
|---|---|
| mci | 0.69 |
| ad | 0.31 |

| Feature | Value |
|---|---|
| Right Ent entorhinal area | 0.44 |
| Right Inf Lateral Ventricle | 0.24 |
| Right FuG fusiform gyrus | -1.70 |
| Right OrIFG orbital part of the inferior frontal gyrus | -2.29 |
| Left Hippocampus | -2.67 |
| Right MTG middle temporal gyrus | -2.06 |
| Right ITG inferior temporal gyrus | -1.75 |
| Right IOG inferior occipital gyrus | -1.58 |
| Left Ent entorhinal area | -0.22 |
| Left Inf Lateral Ventricle | 0.28 |

**Fig. 3**. LIME model output for Support Vector Machines classification of the Mild Cognitive Impairment (MCI) and Alzheimer's disease (AD) classes. False Negative instance, where an AD individual is classified in the MCI class. Left: the predicted probability of the specific instance for the two classes, Right: feature importance ranking and the obtained importance values (coefficients) of the presented features. Orange highlighted features push the prediction to the AD class and features highlighted in blue push the prediction to the MCI class.

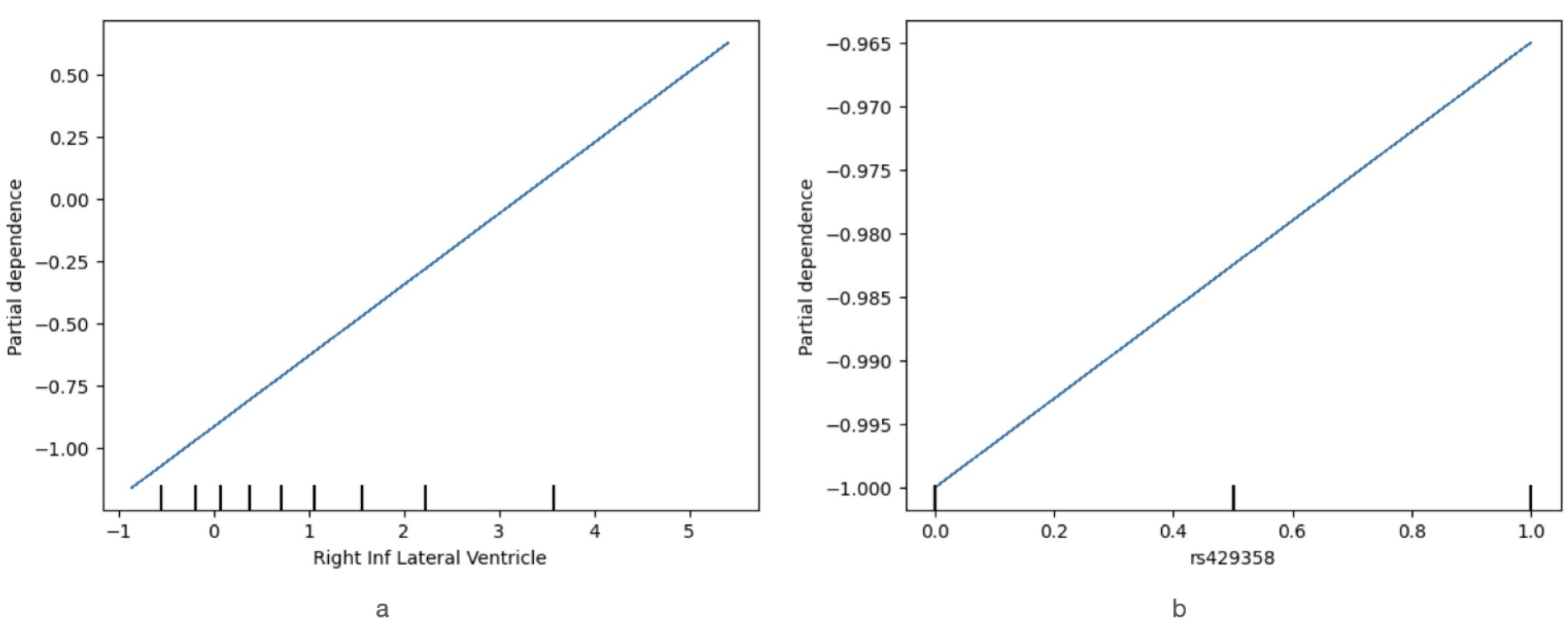

**Fig. 4**. Partial Dependence Plot showing the influence of (**a**) Right Lateral Ventricle and (**b**) rs429358 SNP, which belongs to the ApoE gene[23] in Alzheimer's disease class membership from the classification of Mild Cognitive Impairment (MCI) and Alzheimer's disease (AD) classes, when using the SVM classifier. For the SNP, the value 0.5 in the x-axis is associated with the presence of one allele and value 1 is associated with two alleles.

of one or two alleles (with the value 0.5 in the x-axis being linked to the presence of one allele and value 1 to the presence of two alleles) of the rs429358 SNP, which is associated with the Apolipoprotein E (ApoE) gene[23], almost linearly increased the individual's probability of being classified as an AD rather than an MCI patient, as shown in Fig. 4b.

## Unified explanations of feature attribution-based and counterfactual-based methods

Valuable results were gained by the framework for unifying feature attribution methods with counterfactual explanations, which was used to enhance the robustness of the generated explanations. Unified explanations were produced in the form of feature importance rankings based on the use of Permute Attack[24], as well as through the calculation of the necessity and sufficiency of the considered features, by leveraging SHAP along with counterfactual-based methods, including Permute Attack and Diverse Counterfactual Explanations (DiCE)[25].

To measure feature importance based on the use of counterfactual explanations, the Permute Attack method was applied separately on each binary subproblem, using the SVM-based classifier. In this context, for every pair of classes, one counterfactual example was obtained for every instance in the test set and the number of times each feature was altered in order to contribute to the class overturn, was calculated. The obtained frequency of value alterations for each feature in the generated counterfactuals, considered as an indicator of the feature's importance, is depicted in Fig. 5, where the features are ranked in descending order of frequency in the counterfactual examples, with green (red) bars representing an overall positive (negative) feature change. As most prominent features emerged the ones that were previously highly ranked based on the obtained SHAP feature importance values. In particular, an increase in the patient's lateral ventricles appeared to contribute to the overturn of the MCI prediction and resulted in the individual's classification to the AD class, whereas a decrease in the value of specific features, such as the right entorhinal area, the left middle temporal gyrus or the right amygdala, would result in an MCI patient being classified as an AD patient. Thus, it was inferred that a feature's frequency of occurrence in the counterfactual examples correlated with its importance in distinguishing between the respective classes.

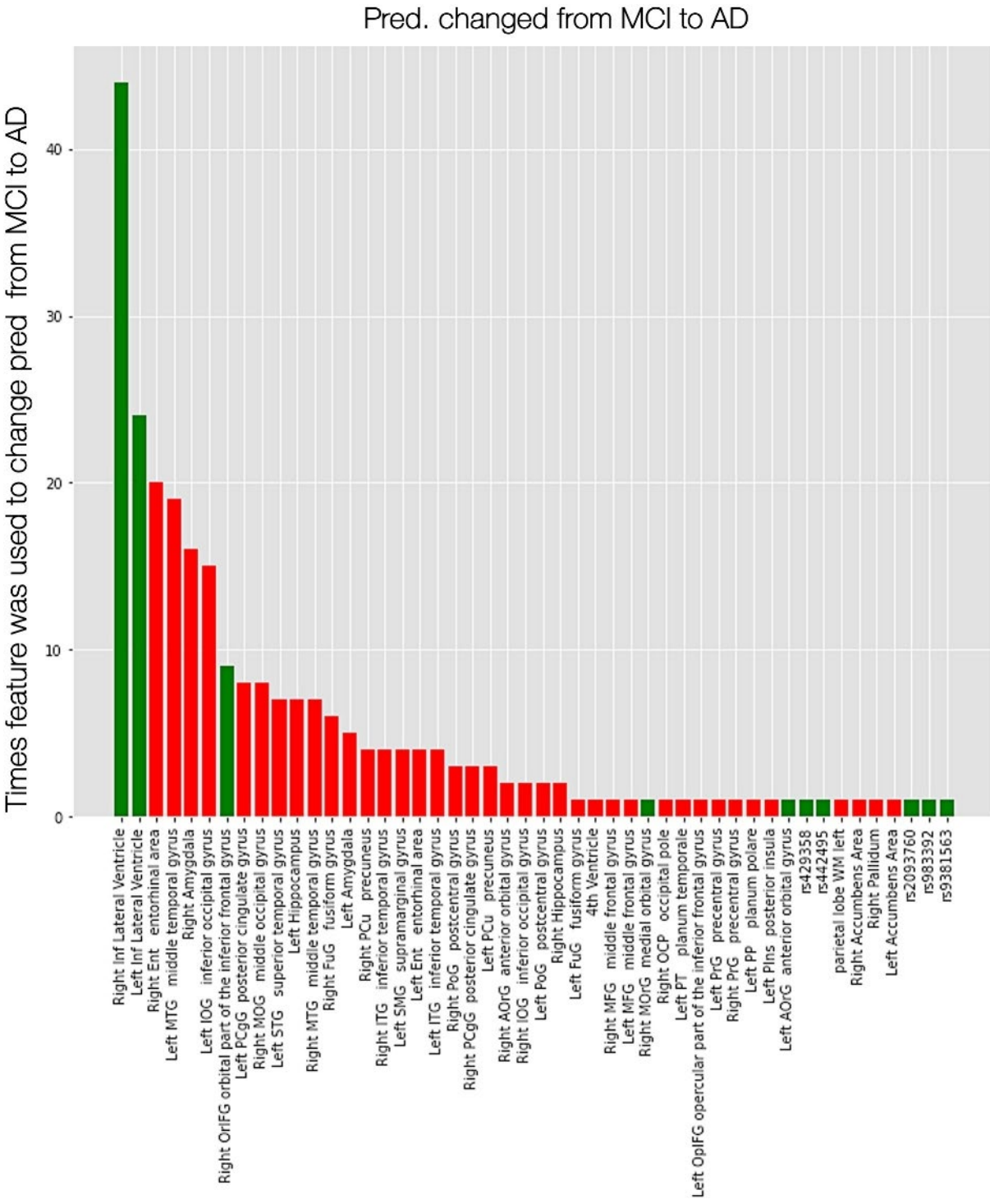

**Fig. 5**. Bar plot showing how many times a feature was selected to be modified in the counterfactual examples, which were created for the test set of the Mild Cognitive Impairment (MCI) and Alzheimer's disease (AD) classes. The features are ordered in descending order of frequency. A green bar represents an overall positive feature value change (increase) and a red bar shows an overall negative feature value change (decrease).

To assess the features' necessity and sufficiency, the top-10 ranked features identified by the SHAP method, which for the MCI versus AD subproblem included, among others, the right and left lateral ventricle, the right and left entorhinal area, and the right and left middle and inferior temporal gyrus, were used to generate counterfactual examples based on the Permute Attack and DiCE methods. The results for necessity and sufficiency were calculated by averaging over 1, 2, 4, and 8 counterfactuals per instance. For the sake of brevity, only the results averaged over 4 counterfactuals per instance are presented. As demonstrated in Fig. 6a, when using Permute Attack, the necessity outcome for each of the top 10 features, as well as for the top 10 features combined, was equal to 0, but when generating counterfactuals by permuting all but the top features, a necessity of 87% was obtained. However, as illustrated in Fig. 6c, when using DiCE for generating counterfactuals, the necessity of the individual top 10 features ranged from 1% to 9%, without a clear correlation between a feature's rank and its corresponding necessity. Notably, the 10th most important feature according to SHAP, the right inferior temporal gyrus, exhibited the highest necessity at 9%, followed by the 5th most important feature, the right middle temporal gyrus, which yielded 7.7% necessity. Moreover, when measuring the necessity of the top 10 features combined, a score of 29.8% was obtained, compared to 63.7% for all but the top 10 features.

As for the top features' sufficiency, Fig. 6 shows that the top 10 features, both individually and collectively, were assigned similar sufficiency values, ranging from 10% to 15% when generating counterfactuals with Permute Attack (Fig. 6b), and from 33% to 38% when using DiCE (Fig. 6d). In the case of Permute Attack, the 9th most important feature, the left entorhinal area, yielded the highest sufficiency (14.9%), whereas with DiCE, the left lateral ventricle, ranked 3rd according to SHAP, achieved the highest sufficiency at 37.9%.

## Discussion

The present study introduced a comprehensive methodological framework aiming at the generation of reliable, interpretable predictions for the diagnosis of MCI or AD. In the classification step, the challenge of multiclass classification and class imbalance, which is related to the imbalanced and multi-stage nature of the patient data related to MCI or AD, was addressed by combining the OVO scheme with the Bagging ensemble learning method. The SVMs, RF and XGBoost classifiers were the highest performing classifiers, while the statistical t-test performed on every pair of classifiers showed no statistically significant differences among the mean balanced accuracy and F1-score of most of the classifiers, with the MLP being the only exception, due to its overall lower performance. The SVM model was selected to demonstrate the application of post-hoc interpretability techniques due to its superior performance with respect to the other classifiers. It should be noted that, unlike inherently interpretable models, in which features contribute in an additive manner to the prediction, SVM's reliance on support vectors and the kernel function, which measures similarity in a higher-dimensional space, makes it a black-box model. Thus, post-hoc interpretability methods were applied towards gaining insights on feature importance.

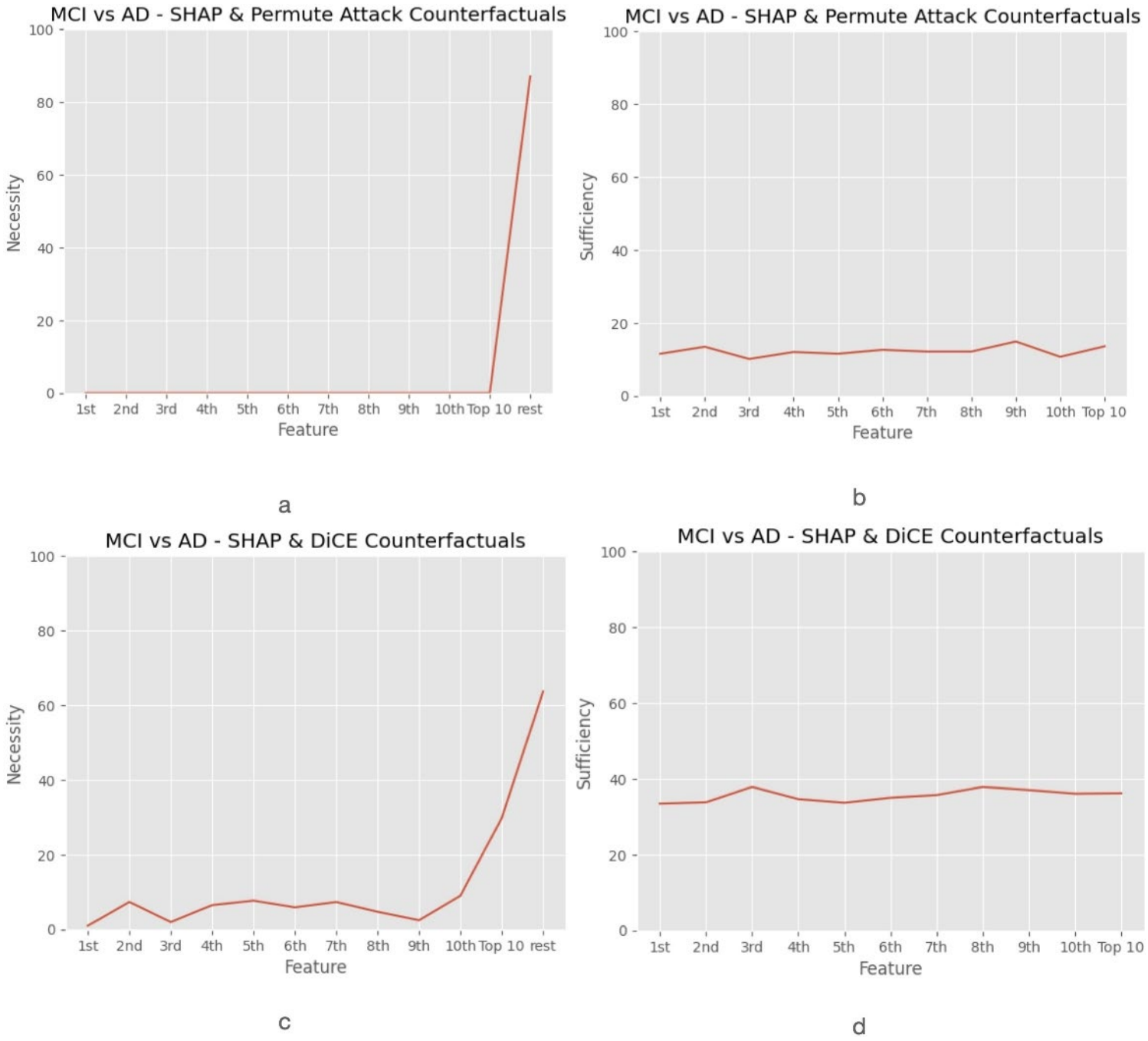

**Fig. 6**. Plot of (**a**), (**c**) necessity and (**b**), (**d**) sufficiency measures of the top-10 features of the SHAP method averaged over 4 counterfactuals per instance using Permute Attack and DiCE methods for generating counterfactuals. For the calculation of necessity, only the k-th most important feature or the top 10 features combined are allowed to change in order to produce counterfactuals, whereas for measuring sufficiency, each time the value of the k-th most important feature or the top 10 features combined are fixed, but all other features are allowed to be permuted for generating counterfactuals.

A comparative assessment between the proposed approach and several recent multiclass classification studies, distinguishing among healthy, MCI and AD participants in the Alzheimer's Disease Neuroimaging Initiative (ADNI) dataset, was performed in terms of classification algorithms, input modalities, model performance, and use of interpretability methods. Although a direct and fair comparison was not feasible due to the application of different input spaces and evaluation frameworks, substantial inferences could be obtained. As shown in Table 2, the discrimination ability of the proposed framework with the best performing classifier (i.e., SVMs) was comparable to that of existing models, even though not all of the considered approaches used the balanced accuracy and the F1-score to assess the models' performance. Notably, the proposed framework's satisfying performance was obtained through leveraging volumetric measurements of MRI brain images in conjunction with genetic data, specifically AD related SNPs, which have rarely been studied together in literature, especially when compared to the more common use of MRI images and clinical data, such as cognitive scores. Moreover, the comparison revealed a lack of comprehensive interpretability methods in prior studies, which were mostly confined to the use of SHAP in more recent works. In contrast, the current work explored the complementarity of various interpretability methods, including SHAP, LIME, and counterfactual explanations, for feature ranking, generating local explanations, and producing counterfactual examples, and harnessed a framework for unifying their results, both in terms of feature ranking and through the assessment of the necessity and sufficiency of highly ranked features, with the ultimate goal to gain an in-depth understanding of the contribution of the various brain regions' volumes and SNPs' presence in the diagnosis of MCI and AD.

To generate rankings of the considered features' importance, the Gini index and the mean absolute SHAP values were utilised. The most prominent ROI features, according to the literature, were highlighted as important both by the Gini index and SHAP. It was observed that smaller values of features, such as the hippocampus, the amygdala, the middle temporal gyri and the entorhinal areas, and the enlargement of the lateral ventricles clearly affected the models' predictions, while these observations were consistent with the relative medical knowledge. Notably, the volume loss of certain anatomical regions of the brain, such as the hippocampus, the amygdala, and the entorhinal areas, is highly linked with the neuropathology of AD[29,30]. Furthermore, the increase in an individual's probability of belonging to either the MCI or AD class in the presence of the enlargement of the lateral ventricles was in accordance with the fact that the cumulative loss of neurons and their connections results in ventricular enlargement, which is strongly correlated with a decline in cognitive performance and is a biomarker of MCI or AD. Ventricular enlargement is primarily linked to grey matter loss and is accompanied by white matter changes due to secondary loss of the axons following neuronal death[30,31]. Gini importance was the only method able to highlight important SNPs features, including rs429358, which is located in the ApoE and is reported to be associated with a higher AD risk[32], rs6448453, which is associated with the CLNK gene and has been shown to increase the risk of developing AD[33], as well as rs2081545 of the MS4A6A gene, which is proven to be linked to the studied diseases according to the literature.

| Study | Features | Number of participants | Classification methods | Achieved performance | eXplainability methods |
|---|---|---|---|---|---|
| Altaf et al. 2018[26] | MRI texture features & Clinical | 287 | SVM, Ensemble, Decision Trees, KNN | 79.8% ACC | None |
| Wang et al. 2018[5] | MRI images | 264 | ensemble of 3D densely connected convolutional networks (3D-DenseNets) | 97.52% ACC | None |
| Basheera et al. 2020[27] | MRI volumetric measurements & Clinical | 120 | CNN | 86.7% ACC | None |
| Lin et al. 2021[28] | MRI volumes, PET intensity, CSF biomarkers, APOE gene | 746 | LDA & ELM | 66.7% ACC, 64.9% F1 | None |
| El-Sappagh et al. 2021[8] | MRI volumetric measurements, PET, Clinical, Genetic | 1043 | **RF**, SVM, GB | 93.33% ACC, 93.82% F1 | SHAP, Decision Trees, FURIA |
| Xu et al. 2022[9] | Clinical | 1074 | SVM-SMOTE & RF | 87.7% F1 | SHAP |
| Yi et a. l 2023[10] | MRI volumetric measurements, APOE gene, Clinical | 1340 | XGBoost-SHAP (XGBoost with adjusted feature weights) | 87.57% ACC | SHAP |
| Amoroso et al. 2023[11] | MRI (brain connectivity network metrics) | 432 | **RF**, SVM, XGBoost, NB, LR | 87.7% F1 | SHAP |
| Current study | MRI volumetric measurements & SNPs | 1463 | RF, LR, MLP, **SVM**, GB, XGBoost | 87.5% balanced ACC, 90.8% F1 | Gini index, SHAP, LIME, PDP, Counterfactuals & unification |

**Table 2.** Comparison of different studies for CN/MCI/AD multiclass classification in the ADNI dataset, based on input features, number of participants, classification methods, performance and explainability techniques. Where multiple classifiers were evaluated, the ones in bold are the ones which yielded the highest performance. LDA: Linear Discriminant Analysis, ELM: Extreme Learning Machine, FURIA: Fuzzy Unordered Rule Induction Algorithm, ACC: Accuracy, F1: F1-Score.

PDPs were deployed to capture how each individual feature influenced the prediction on average. This approach confirmed some of the previous observations regarding the effect of features' values, such as the smaller or larger volume of certain brain regions, but it also provided valuable insights about the role of specific SNPs features. For instance, the presence of one or two alleles of the rs429358 SNPs, which was highlighted among the highly ranked features by the Gini index, was found to increase the AD risk, while the presence of one or two alleles of the CASS4 gene[34] or the MS4A6A[35] gene related SNPs was associated with a smaller chance of MCI or AD (as shown in the Supplementary material). This is confirmed by the fact that these genes have been linked to lower susceptibility to certain neurodegenerative diseases and play a protective role against brain atrophy.

The generation of local explanations was also investigated with the use of LIME to highlight how features' values affected individual predictions. LIME allowed the comparison of the local feature importance based on different classifiers for the same dataset instances. In most cases, the obtained rankings of the most impactful features were slightly differentiated among instances and between models. It was also observed that even when the classifiers' predictions were not in agreement (example in the Supplementary material—Figs. S7 and S8), the features' values had a similar impact on the prediction probabilities.

The outcomes of global and local feature attribution by Gini, SHAP, LIME, and feature importance rankings based on counterfactuals were largely consistent in terms of highly ranked features. These methods consistently identified common attributes, including the entorhinal areas, the lateral ventricles, the middle temporal gyri and the right fusiform gyrus. The observed differences in the highlighted features among XAI methods stemmed from fundamental distinctions in each approach's operating principles and perspective of assessing feature importance (i.e., providing global or local explanations). The robustness of the complementary interpretability outcomes derived from SHAP feature attribution and counterfactual-based methods was assessed by the calculation of necessity and sufficiency, which provided supplementary quantifiable measures of the individual role of highly ranked features in shaping model predictions.

Regarding the framework for unifying feature attribution methods and counterfactual explanations, the low values of necessity obtained by Permute Attack and DiCE indicated that the SHAP attribution method could not ensure high necessity for any of the top-10 ranked features alone. For both counterfactual methods, the high necessity of all other features combined reflected the necessity of all those features for the model output. Notably, in the case of DiCE, the necessity of the top 10 features combined (i.e., 29%), which corresponded to the top 5% features of the dataset, was nearly half the combined necessity of the remaining 95% of features (i.e., 63%), thus highlighting the relatively high necessity of the combination of the top 10 features, especially considering the high number of features in the dataset. The obtained sufficiency measures demonstrated that each of the top features alone, as well as the top 10 features combined, were to some extent sufficient for the model output, despite the very large amount of features in the dataset. Both Permute Attack and DiCE resulted in no significant variations across the sufficiency scores of the individual or combined top features, meaning that each of the top features alone was considered sufficient for the model predictions to the same extent as the combination of the top 10 features.

In general, the top 10 features' necessity and sufficiency values were expected and confirmed to present small differentiations, due to the similar mean Shapley values attributed to the top features according to the SHAP ranking. However, discrepancies were observed between feature rankings based on Shapley values and the corresponding necessity scores, which can be attributed to the definition of necessity, involving complementary interpretability aspects captured by the SHAP analysis and counterfactual explanations. While Shapley values quantify a feature's marginal contribution to the model's output, necessity measures rely on counterfactual generation, which is constrained by the ability to produce valid counterfactuals when perturbing only one feature at a time. Moreover, the large number of features in the dataset was linked to the low values of necessity, since highly ranked features may often neither be necessary nor sufficient and these properties are bound to decrease as the number of features in a dataset increases[18]. The observed differences in the obtained necessity and sufficiency results between Permute Attack and DiCE could be related to the methods' distinct approaches to generating counterfactuals, particularly regarding diversity and proximity. Permute Attack focuses on maintaining proximity in the generated counterfactuals, prioritising minimal changes to the original feature values[24]. This proximity constraint may justify the method's inability to produce counterfactuals when restricted to modifying one feature at a time, thus leading to the obtained zero values of necessity, as changing the model's output would possibly require a significant change in the respective feature's value. On the other hand, DiCE emphasises on striking a balance between diversity and proximity, enabling it to produce counterfactuals even with single feature modifications towards calculating the necessity score. This indicates that the calculation of necessity and sufficiency is highly dependent on the selected counterfactual generation method's capacity to create unique and valid counterfactual examples.

Potential limitations of the proposed framework may be associated with the fact that the applied interpretability methods and their unification can only provide results pertaining to each pair of classes separately, rather than to all problem classes together. Another limitation is related to the sensitivity of the unification framework to the counterfactuals generation method, which may affect the coherence of the obtained results in terms of necessity and sufficiency. To further improve the applicability and generalisability of the proposed framework, future work includes extending the evaluation of the proposed framework on data from multiple cohorts. Moreover, the exploration of multiclass interpretability techniques, in order to provide explainable outputs for the final prediction from all classes, will be considered, and the use of additional classification models, such as deep learning algorithms, will be investigated to further improve the classification accuracy of the framework. The inclusion of additional modalities, such as cognitive assessments and the evaluation of their contribution to the models' performance will also be considered. The exploration of further techniques for unifying the results of the interpretability methods can be investigated to highlight other important aspects of AD neuroanatomical changes.

## Methods

### Overview of the conceptual framework

This proposed framework aims at providing robust interpretable predictions of MCI or AD diagnosis, while addressing key challenges related to data heterogeneity, multiclass imbalance, reliability, and transparency. It entails an ensemble classification method combined with a multiclass classification scheme and a nested cross validation scheme for the performance evaluation and hyperparameter tuning of the tested classifiers. The use of various interpretability methods, including feature attribution techniques and counterfactual explanations, is investigated for generating human-friendly interpretations on the outputs of the best performing classifiers. A unification approach is subsequently deployed for combining the aforementioned interpretability methods and assessing the considered features' necessity and sufficiency. A schematic representation of the proposed framework is provided in Fig. 1.

### Dataset

In order to assess the ability of the proposed framework to provide reliable, robust, and interpretable predictions for MCI or AD diagnosis, a medical cross-sectional dataset of 1463 subjects aged between 60 and 86 years old, obtained from the ADNI database (adni.loni.usc.edu) was utilised. The dataset included baseline data from 449 healthy controls CN, 740 patients with MCI, and 274 AD patients. The considered features consisted of demographic data and clinical data, comprising 154 volumetric ROIs obtained from T1-weighted MRI brain scans, as well as of 54 AD related single nucleotide polymorphisms (SNPs), which described whether the participant carried zero, one, or two alleles. The studied problem constituted a multiclass classification problem, characterised by class imbalance, since the MCI class contained almost half of the total dataset instances. Information about the demographic characteristics of participants from each class can be found in the Supplementary Material.

The scan of each participant was first corrected for intensity inhomogeneities and a multi-atlas skull stripping algorithm, specifically a multi-atlas label fusion method, was implemented for the removal of extra-cranial material[36,37]. To obtain the ROI features, brain tissue segmentation was performed using a multi-atlas segmentation technique[38]. Subsequently, a linear covariates adjustment was applied, aiming to remove age, sex, and brain volume effects and retain the disease-associated neuroanatomical variation of the ROI volumes. Age, sex, and the total brain volume covariates' correlation with the ROI volumes of the 449 healthy controls was predicted with Linear Regression and all ROI features were respectively residualised. Afterwards, all ROI volumes were standardised by using the controls' mean and standard deviation values to calculate the z-score measure.

## Classification space

The classification step of the proposed framework focused on combining methods, which are applicable both to multiclass problems as well as to problems with a significant class imbalance, and also investigated and evaluated the use of a variety of classifiers, including SVMs, tree-based algorithms, such as RF, GB, and XGBoost, and deep learning models such as MLP. In case of a multiclass classification problem, the most common approach is to decompose it into several binary subproblems and then apply a voting scheme to make the final class prediction from the outputs of the binary classifiers. The two prevalent decomposition approaches are the OVA and the OVO methods[39], both of which were explored within the proposed framework. For an m-class dataset, OVO requires decomposition into $\frac{m\cdot(m-1)}{2}$ binary problems, considering every possible combination of classes, while OVA generates m distinct binary subproblems, each one including one of the original classes and a class featuring all the samples that belong to the remaining $m-1$ original classes.

In order to tackle the problem of class imbalance, an ensemble learning strategy based on the Bagging method, also known as Bootstrap and Aggregating, was deployed. This technique was selected due to its ability to successfully address class imbalance and at the same time reduce the individual predictions' variance, thus enhancing the performance and robustness of the model[41,42]. In this context, the MCI instances of the training set, constituting the majority class, were split into two fractions and two different training subsets were created, both containing all CN and AD training samples and each of them a different half of the majority class samples. The OVO method was applied to both distinct training subsets, since it performed better than the OVA approach. For each binary subproblem, two primary classifiers were trained on each training subset. The OVO voting strategy was used to combine the outputs of the primary classifiers, thus attributing each test set instance to the class with the maximum predicted probability. With the same voting strategy, the aggregation step of Bagging ensured the final prediction for all instances of the test set. To find the optimal hyperparameters of the various classifiers and ultimately evaluate and compare their performance, a 5x4 fold nested cross validation scheme was applied and the final evaluation was based on the balanced accuracy metric, selected due to its ability to account for the existing class imbalance.

## Generation of feature attributions and counterfactual explanations

Various interpretability methods were utilised in order to provide explanations on the classification predictions, as well as the importance and attribution of the considered features. A framework based on actual causality was subsequently harnessed, aiming to unify feature attribution-based explanations using counterfactual examples.

The SHAP[20] method was deployed due to its ability to produce global explanations, reflecting features' contributions to the model's outputs, and, thus, highlight key aspects of the model's decision process across all instances in the form of feature ranking[6,40]. Since tree-based classifiers were also explored for the diagnosis of MCI and AD, the Gini Importance global feature attribution measure was also calculated[19]. In order to dive into the local feature attribution, the LIME[21] method was applied on individual predictions to measure the features' influence on the corresponding class probabilities[19]. Moreover, PDPs' ability to produce visualisations of the global influence of a specific feature to the final prediction motivated their use towards separately examining the importance of individual features[22]. Counterfactual explanations and, specifically, the Permute Attack[24] and DiCE[25] methods, were applied in the context of the unification framework to produce measures of the considered features' importance based on the calculation of the fraction of times a feature was modified to alter the predicted model outcome. Moreover, counterfactual explanations were combined with feature attribution methods towards measuring the necessity and sufficiency for top ranked features.

## Unifying feature attribution-based methods and counterfactual explanations

To generate unified explanations of attribution-based methods and counterfactuals, two distinct approaches were applied. The first approach produces feature attributions based on the use of counterfactual examples and relies on the notion that important features are more likely to be permuted when generating counterfactuals compared to less important ones. An attribution score is, thus, calculated by determining the fraction of times a feature's value is modified during counterfactual examples' generation. To measure local feature attribution, this score is averaged over multiple counterfactual examples generated for a single instance in the dataset, whereas for a global feature attribution score, the number of times a feature was permuted is averaged over multiple test instances.

The second approach, grounded in actual causality, provides unified explanations by harnessing the complementarity of feature attribution methods and counterfactual explanations towards the generation of reliable model explanations. It aims at evaluating feature importance by examining the necessity and sufficiency of highly ranked features in contributing to the model's predictions, thus assessing how indispensable a feature is (necessity) and whether it alone can guarantee the outcome (sufficiency). According to the deployed approach, assuming that a subset of feature values $x_j = a$ is an explanation for a model output $y^*$ relative to a set of contexts *U*, the notion of but-for causes captures the necessity of a particular feature subset for the obtained model output, namely for each context $u \in U$, where $x_j = a$ and $f(x_{-j} = b, x_j = a) = y^*$ and a subset of features $x_{sub} \subseteq x_j$ is an actual cause under a specific configuration of the input *x* and the output *f*(*x*) of the model. Similarly, sufficiency means that setting a feature subset $x_j \leftarrow a$ will always lead to the given model output, irrespective of the values of other features, so the definition of sufficiency is that for all contexts $u' \in U$ , $x_j \leftarrow a \Rightarrow y = y^*$[18].

The metric of necessity aims to measure how necessary a subset of feature values is for the model's output and for a specific feature value it is calculated by only allowing this feature to change while generating counterfactuals and finding the fraction of times that changing this feature value leads to a valid counterfactual, thus indicating the extent to which this was necessary for the model's output. Necessity is described as the probability that feature $x_j$ is a cause of output $y^*$, given that $x_j = a$ and $y = y^*$. Let $y^* = f(x_j = a, x_{-j} = b)$ be the output of classifier *f* for input *x* and $x_j = a$ a feature value. To measure the necessity of $x_j = a$, counterfactuals are generated by only allowing $x_j$ to be changed and the fraction of times that valid counterfactuals are produced is calculated. If the model's output is altered by modifying $x_j$, then $x_j = a$ is necessary to generate the original output. The definition of necessity is:

$$\text{Necessity} = \frac{\sum_{i, x_j \neq a} \mathbb{1}(CF_i)}{nCF * N}, \tag{1}$$

where *nCF* represents the number of counterfactual examples to be generated for each instance and *N* is the total number of instances for which counterfactuals are produced.

The metric of sufficiency aims to calculate how sufficient a given subset of feature values is for the model's output and it is measured by generating counterfactuals with all but a specific feature and calculating the fraction of the times those unique counterfactuals are generated subtracted from 1, thus, the fewer the generated unique valid counterfactuals, the more sufficient the feature. Sufficiency is calculated using the conditional probability of the output $y = y^*$ given that $x_j \leftarrow a$. More specifically, to measure sufficiency, $x_j$ is fixed to its original value and all other features are allowed to change when generating counterfactuals. If valid counterfactuals are not produced, then $x_j = a$ is sufficient for causing the model's output, else, (1 - the fraction of times these unique counterfactuals are generated) indicates the extent of sufficiency of $x_j = a$. Sufficiency is therefore measured by subtracting the number of unique counterfactuals generated by keeping $x_j$ fixed, from the fraction of unique counterfactuals produced by allowing all the features to change and is defined as follows:

$$\text{Sufficiency} = \frac{\sum_i \mathbb{1}(CF_i)}{nCF * N} - \frac{\sum_{i, x_j \leftarrow a} \mathbb{1}(CF_i)}{nCF * N}, \tag{2}$$

where *nCF* is the number of counterfactual examples to be generated for each instance and *N* represents the total number of instances[18].

## Data availability

Data used in the preparation of this article were obtained from the Alzheimer's Disease Neuroimaging Initiative (ADNI) database. The ADNI data base is public for researchers and can be downloaded upon request at (https://adni.loni.usc.edu.)

## Code availability

The code for reproducing the experiments of this study is available at the following Github repository: https://github.com/Marily-Vlontzou/XAI-framework-for-MCI-AD-diagnosis.

## Acknowledgements

Data used in the preparation of this article were obtained from the Alzheimer's Disease Neuroimaging Initiative (ADNI) database (adni.loni.usc.edu). The ADNI was launched in 2003 as a public-private partnership, led by Principal Investigator Michael W. Weiner, MD. The primary goal of ADNI has been to test whether serial magnetic resonance imaging (MRI), positron emission tomography (PET), other biological markers, and clinical and neuropsychological assessment can be combined to measure the progression of mild cognitive impairment (MCI) and early Alzheimer's disease (AD). For up-to-date information, see (www.adni-info.org.) Data collection and sharing for this project was funded by the Alzheimer's Disease Neuroimaging Initiative (ADNI) (National Institutes of Health Grant U01 AG024904) and DOD ADNI (Department of Defense award number W81X-WH-12-2-0012). ADNI is funded by the National Institute on Aging, the National Institute of Biomedical Imaging and Bioengineering, and through generous contributions from the following: AbbVie, Alzheimer's Association; Alzheimer's Drug Discovery Foundation; Araclon Biotech; BioClinica, Inc.; Biogen; Bristol-Myers Squibb Company; CereSpir, Inc.; Cogstate; Eisai Inc.; Elan Pharmaceuticals, Inc.; Eli Lilly and Company; EuroImmun; F. Hoffmann-La Roche Ltd and its affiliated company Genentech, Inc.; Fujirebio; GE Healthcare; IXICO Ltd.; Janssen Alzheimer Immunotherapy Research & Development, LLC.; Johnson & Johnson Pharmaceutical Research & Development LLC.; Lumosity; Lundbeck; Merck & Co., Inc.; Meso Scale Diagnostics, LLC.; NeuroRx Research; Neurotrack Technologies; Novartis Pharmaceuticals Corporation; Pfizer Inc.; Piramal Imaging; Servier; Takeda Pharmaceutical Company; and Transition Therapeutics. The Canadian Institutes of Health Research is providing funds to support ADNI clinical sites in Canada. Private sector contributions are facilitated by the Foundation for the National Institutes of Health (www.fnih.org). The grantee organization is the Northern California Institute for Research and Education, and the study is coordinated by the Alzheimer's Therapeutic Research Institute at the University of Southern California. ADNI data are disseminated by the Laboratory for Neuro Imaging at the University of Southern California. The investigators within the ADNI contributed to the design and implementation of ADNI and/or provided data but did not participate in analysis or writing of this report. A complete listing of ADNI investigators can be found at: https://adni.loni.usc.edu/wp-content/uploads/how_to_apply/ADNI_Acknowledgement_List.pdf.

## Author contributions

M.E.V performed the analysis, implemented the methodology, performed the experiments and wrote the article. K.D and I.S co-advised during the methodology design phase, performed literature review, helped analyse the results and reviewed the article. M.A. also advised during the methodology conceptualisation phase, had a critical role in the manuscript revision process, providing extensive feedback during multiple rounds of revisions and helped edit and format the article, thus significantly improving the clarity and quality of the paper. C.D. preprocessed the data, supervised the process, and provided valuable feedback. K.N. supervised the process and provided critical feedback throughout the research and manuscript preparation stages. All authors reviewed the manuscript and approved the submitted version.

## Competing interests

The authors declare no competing interests.

## Additional information

**Supplementary Information** The online version contains supplementary material available at https://doi.org/10.1038/s41598-025-92577-6.

**Correspondence** and requests for materials should be addressed to M.E.V.